# An IIoT machine model for achieving consistency in product quality in manufacturing plants

Abhik Banerjee, Abdur Rahim Mohammad Forkan, Dimitrios Georgakopoulos, Josip Karabotic Milovac, Prem Prakash Jayaraman
*Department of Computing Technologie, School of Science, Computing and Engineering Technology*
Swinburne University of Technology
Melbourne, Australia
{abanerjee, fforkan, dgeorgakopoulos, jkaraboticmilovac, pjayaraman}@swin.edu.au

*Abstract*— Consistency in product quality is of critical importance in manufacturing. However, achieving a target product quality typically involves balancing a large number of manufacturing attributes. Existing manufacturing practices for dealing with such complexity are driven largely based on human knowledge and experience. The prevalence of manual intervention makes it difficult to perfect manufacturing practices, underscoring the need for a data-driven solution. In this paper, we present an Industrial Internet of Things (IIoT) machine model which enables effective monitoring and control of plant machinery so as to achieve consistency in product quality. We present algorithms that can provide product quality prediction during production, and provide recommendations for machine control. Subsequently, we perform an experimental evaluation of the proposed solution using real data captured from a food processing plant. We show that the proposed algorithms can be used to predict product quality with a high degree of accuracy, thereby enabling effective production monitoring and control.

## I. Introduction

Ensuring consistency is key to improving the production efficiency and productivity of manufacturing plants. Variability in production leads to increased waste and longer production runs, which significantly reduce the productivity of manufacturing plants [1]. However, achieving consistency is challenging as it involves coordination among all manufacturing systems, including machinery and processes.

In particular, ensuring consistency of product quality is a challenging area for manufacturing plants, as it requires balancing a large number of manufacturing attributes. Product quality measurements are typically available quite late in the production process and deviation from target product quality leads to increased wastage and reprocessing, thereby inhibiting plant productivity. With the objective of minimizing variability, existing production practices typically involve frequent adjustments to plant machinery during production runs, which requires continuous and close monitoring of the machines based on machine sensor data, which is difficult due to multiple reasons. Firstly, a large number of machine sensors are difficult to track individually. Secondly, it is difficult to determine the impact of combinations of multiple machine sensor data on the product quality. This is further aggravated by the fact that machine sensor values change continuously during production, making it difficult to identify critical states of machine operation. Finally, determining which machine settings are to be used, and how they are to be adjusted, to control machinery is also a difficult challenge for similar reasons. Effective machine control involves achieving the right combination of multiple machine settings, which is made difficult due to a large number of machine settings, and because interdependence among machine settings, machine sensor, process and product quality data is not well understood. As a result of all the above challenges, existing practices are driven mostly by intuition and experience, which is often in itself a cause for variability [2]. Thus, in order to improve consistency of product quality, there is a need for a solution that enables the following:

- Integration of machine data with process and product quality data

- Determination of critical states of machine operation, with regard to their impact on product quality

- Estimation of product quality based on machine data

- Recommendations for machine settings adjustments during production to achieve the target product quality, thereby achieving consistency

In this paper, we propose a data-driven approach for achieving consistency in product quality. In particular, we propose an IIoT machine model for monitoring machines in real-time. The IIoT machine model can be used to monitor machine operations during production runs, by generating snapshots of machine operations during production, as well as how these snapshots are related to the final product quality. Subsequently, we present algorithms for performing predictive and prescriptive analytics using the IIoT machine model, to predict product quality in real-time as well as provide machine settings recommendations.

In summary, the key contributions of this paper are:

a) A novel IIoT machine model, which provides an abstraction of the machine operations in real-time.

b) Algorithms for performing predictive and prescriptive analytics, which use the IIoT machine model to provide product quality predictions and machine settings recommendations in real-time.

c) Experimental evaluation of the IIoT machine model, and the predictive and prescriptive analytics algorithms, using manufacturing data obtained from a real-world food processing use-case.

## II. Related work

Industrial Internet of Things (IIoT) [3] is expected to become a general norm in manufacturing plants [4]. The availability and use of low-cost Internet of Things (IoT) sensors have enabled generation of large amounts of data in manufacturing plants [5] pertaining to machines, processes and products. Effective data analyses that incorporate all this data, as well as the relationships among them, is necessary to derive meaningful insights.

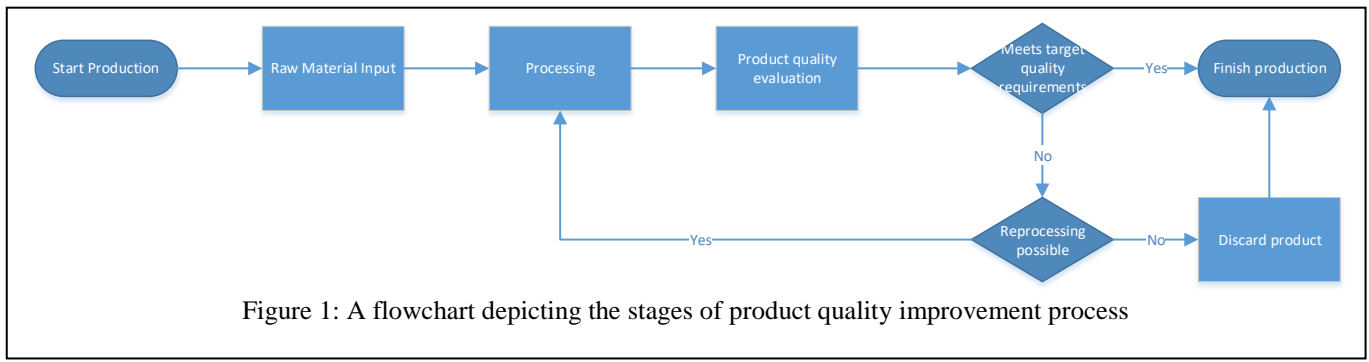

Figure 1: A flowchart depicting the stages of product quality improvement process

Regular operations in manufacturing industries are carried out by operators and rely on expert intuition and experience, and pre-defined rules to resolve operational issues during production lifecycle [6]. However, the effectiveness of such human-centric processes is limited in a dynamic manufacturing environment where the type of data generated from heterogeneous machines and processes can be continuous and have a high degree of variations [7]. This is because production outcomes such as product quality are often impacted by small changes in machine settings, thereby impacting the consistency in product quality. In a fast-paced manufacturing process operators are required to make quick judgements about machine settings to ensure consistency in product quality [8]. The need for accurate decision making has necessitated the move towards data-driven solutions [5].

Machine learning (ML) models have previously been applied to handle several areas in manufacturing such as predictive maintenance [9] and fault detection [10], which are typically addressed by applying ML on localized data pertaining to the problem, , whereas other applications such as process optimization take into account data pertaining to the entire manufacturing process [11]. However, production outcomes are dependent on not just machine configuration but also on how machines are operated during production. This makes it difficult to determine beforehand how subsets of high-dimensional machine data are correlated with individual production outcomes [12]. This necessitates the need to model a machine as a whole, so as to enable dynamic monitoring of machine operational states, including the impact of control actions such as changes in machine settings. In [13], the authors propose a technique of modeling assets in order to determine the optimal control actions to maintain the optimal asset condition. While this model assumes fixed asset states, the same is not true for machines, as the importance of operational states of machines is dependent on the production outcome which is to be monitored, such as product quality. While a popular approach to identify preliminary relationships between data is to use logistic regression [14], such regression based methods become ineffective with the increase in the number of variables, as with machine data. In this paper, we present interpretable machine models which are then used to generate predictions and recommendations in real-time.

III. PRODUCT QUALITY MONITORING AND CONTROL IN MANUFACTURING

In this section, we describe the stages of manufacturing process that deal with monitoring and controlling of product quality. We first describe how product quality monitoring and control is performed in different manufacturing domains, subsequently present a generalization of this process, and finally describe the manufacturing data associated with it.

A. *Product quality monitoring and control in different manufacturing domains*

We describe in brief the product quality monitoring and control process in three different manufacturing domains, but see that they share similarities, which can be generalized.

*1) Food processing*

Ensuring consistency in quality of food products is critical as it impacts both food safety and taste. Some examples of food processing include production of food pastes [20], jams and jellies [21] and involve boiling, drying or evaporation to obtain the final product. Here, the product quality is typically measured in terms of the total soluble solids, with the acceptable quality varying depending on the type of food product. For instance, the target level of total soluble solids for jams is between 65 - 68%. The production process requires continuous monitoring and controlling of the evaporator/boiler/dryer machines to achieve the target quality. The large number of machine sensors and machine settings contribute to the complexity of such processes, because of which existing practices rely on manual estimation [22].

*2) Manufacturing of composites*

Carbon composites are used in a wide range of industries, such as in manufacturing of automotive components. Despite their attractiveness as a material due to lightness and strength, manufacturing of carbon composites is costly, which makes it critical to perform continuous quality monitoring [23]. As with food production, achieving the target product quality requires tuning multiple machine attributes, along with the processing time. The product quality is determined as the level of cross-linking of fibers, which is estimated using pressure, temperature and dielectric sensors (that measure ion viscosity). Upon detection of a fault, the product needs to be discarded. This makes it necessary to determine critical control points, and adapt process parameters to minimize wastage.

*3) Steel manufacturing*

Steel manufacturing involves converting raw materials into steel products such as steel billets [24]. However, the manufacturing process often results in defects such as cracks or ripples on the surface of the manufactured product or deformations in shape. Achieving consistency in the quality of manufactured steel products requires effective control of machines during steel production.

### B. Generalized product quality monitoring and control process

Based on the above examples, we now describe a generalized product quality monitoring and control process as applied to any manufacturing domain. Irrespective of the manufacturing domain, a product quality control and monitoring process can be divided into four main stages.

*1) Raw material input*

Before starting production, the first step is to provide the raw materials as input. At this stage, the raw materials may be put into storage from where it can be picked up by plant machinery for processing. The raw material quality may be measured to verify that it matches the production requirements, failing which it may either be discarded or made to undergo pre-processing.

*2) Production setup*

The production setup stage involves setting up the machines for a production run. This typically involves configuring machines with pre-defined settings. The machine settings for setup are typically determined theoretically as those that are most likely to achieve the target product quality. However, these settings do not adapt to variations in production, which could arise from multiple factors such as raw material variations, seasonal variations, changes to production environment, etc.

*3) Processing*

The processing stage involves manufacturing of the product by processing the raw material using the plant machinery. This stage requires continuous monitoring of machine sensors and control through machine settings adjustments. Machine settings adjustments are often done based on tacit knowledge and human experience, and sometimes based on pre-defined rules.

*4) Quality evaluation*

The final stage involves measurement of the product quality using one or more sensors, which may be online, wherein quality measurement sensors are a part of plant machinery and provide real-time measurements, or offline, requiring manual testing with a separate quality measurement device using the product. If quality measurements are not found to meet the desired targets, it may either lead to reprocessing of the product, or require discarding the product.

Figure 1 depicts the product quality monitoring and control for an individual production run.

### C. Manufacturing data pertaining to product quality monitoring and control process

Data pertaining to different aspects of the product quality monitoring and control process can be categorized as follows:

1. **Process data:** Data pertaining to the product quality monitoring and control process are recorded using *process control software*, and includes the following:

    a. ***Production batch data:*** A production batch refers to a group of products produced together. All product units that are part of the same production batch are identified together, such as a batch id.

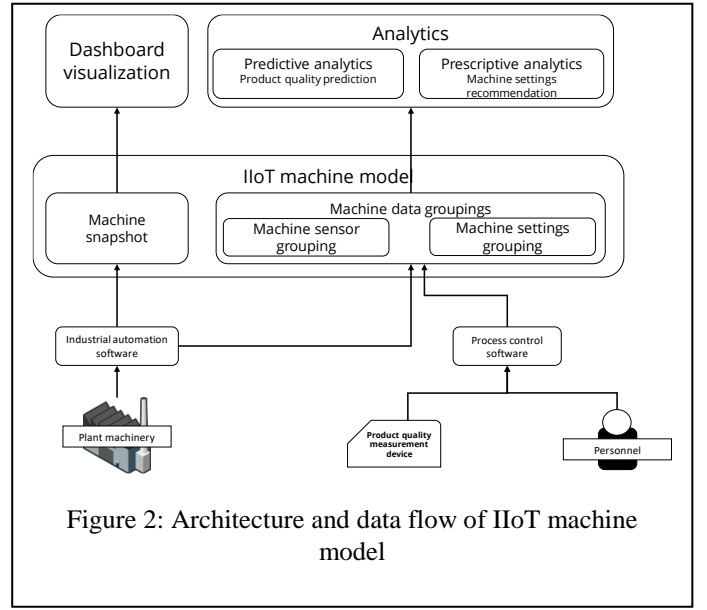

Figure 2: Architecture and data flow of IIoT machine model

   b. ***Production run:*** A production run refers to contiguous manufacturing of products on the production line. A production run $r$ has a unique identifier to distinguish it from other production runs. For the discussion in this paper, we consider that an individual production run corresponds to a single production batch, and the batch id uniquely identifies the production run. The production run data consists of the batch id and the time duration characterizing the production run. Thus, a production run $r$ is characterized as $r = \{B_r, t_s^r, t_e^r\}$, where $B_r$ is the batch id, while $t_s^r$ and $t_e^r$ are the start and end times of the production run $r$.

2. **Product quality data:** Data pertaining to the product quality of production runs, and consists of the following:

    a. ***Production run quality:*** Refers to the quality of the product manufactured in a production run $r$. The production run quality is denoted as a quality label $q_r \in Q_L$, where $Q_L$ is the set of possible quality labels for a manufacturing process. The quality label for a single production run may be derived from multiple quality measurements taken during the production run.

    b. ***Target product quality:*** Refers to the optimal product quality $\widehat{q_r}$ .which is targeted for a production run

Product quality data is typically recorded along with production data in *process control software.*

3. **Machine data:** Includes all machine sensor and machine settings data recorded during production runs, Machine data is typically captured using *industrial automation software.*

### D. Objectives of a data-driven solution

For the product quality monitoring and control process described above, we now identify the objectives of a data-driven solution, in order to improve consistency in product quality:

1. **Machine monitoring:** Effective monitoring of machines is necessary to determine subsequent actions, which may include machine control actions such as changing machine settings or obtain offline product quality measurements. Hence, a data-driven approach needs to enable obtaining the following information:
    a. *Machine snapshot:* It is necessary to have a complete view of machine operations at any time instant during a production run, in order to estimate subsequent outcomes and to decide on control actions. We term this as the machine snapshot, which needs to include the instantaneous machine sensor data, along with the applied machine settings. Further, the machine snapshot also needs to include the most recent changes made to machine settings, as this can help estimate how the machine will behave in subsequent time instances. For instance, if the applied machine setting of temperature is at 80 °C but the most recent change was a temperature reduction of 10 °C, it would imply that the machine is expected to undergo some cooling, which may also impact other machine sensor parameters.
    b. *Impact of machine snapshot on product quality:* In addition to the machine snapshot, it is necessary to determine how the machine snapshot relates to the production outcome, i.e., the product quality. Information about the estimated product quality that machine snapshots are likely to lead to can help determine subsequent machine control actions.
2. **Machine control:** In order to enable optimal machine control during production, the data-driven solution needs to provide the following:
    a. *Product quality estimation:* Estimation of the product quality achievable for machine control actions at different stages of production can help choose the machine control actions, both during production setup and during processing.
    b. *Machine control recommendations:* While product quality estimation is predictive in nature, prescriptive analytics outcomes [25] can be provided in the form of recommendations of machine control actions to be performed at setup and during production. Such recommendations could be directly applied to the machine.

IV. IIoT MACHINE MODEL

We propose an IIoT machine model which can be used to monitor machine operations. The proposed IIoT machine model includes (a) machine snapshot that provides a comprehensive view of machine operations at any time instant, (b) machine sensor states that characterize how different combinations of multiple machine sensors are related to production outcomes, and (c) machine settings states characterizing how combinations of machine settings relate to production outcomes.

*A. Architecture and data model*

The proposed IIoT machine model uses instantaneous and historical production data to characterize the machine operation with regard to product quality outcomes. Figure 2 shows how the architecture for generating the IIoT machine model using manufacturing data from plant machinery, which are obtained from existing industrial automation software and process control systems.

The IIoT machine model can be used by other applications to enable efficient machine monitoring through efficient visualization and through predictive and prescriptive data analytics. In section VI, we describe predictive and prescriptive analytics techniques which use the IIoT machine model to predict product quality outcomes of a production run. Later, in section VII, we evaluate it using a real-world experimental use case.

The data model for the IIoT machine model is depicted in Figure 3 as an Entity-Relationship diagram. The data model consists of the following entity types:

1. The *machine sensor* and *machine settings parameters* include the list of machine sensors and machine settings which are used in the IIoT machine model.
2. The *machine sensor* and *machine settings data* recorded at individual time instances during production runs.
3. The *machine snapshot* data includes all operational states of the machine consisting of machine sensor as well as machine settings data. This is detailed further in section IV.B.
4. The *machine sensor* and *machine settings states* include all states that characterize the machine operation with respect to product quality. The states are generated by taking as input the historical machine snapshot data as well as product quality data. The generation of machine sensor and machine settings states is detailed in section IV.C.

*B. Machine snapshot*

The machine snapshot consists of instantaneous machine data, $d(t)$ at any time instant $t$, which includes the following:

- *Machine sensor observations:* Includes all the machine sensor values at any time instant $t$ and can be represented as $s(t) = \{s_1(t), \ldots, s_{n_s}(t)\}$, where $n_s$ is the count of machine sensors
- *Machine settings observations:* Includes the values of all machine settings at any time instant $t$ and can be represented as $h(t) = \{h(t), \ldots, h_{n_h}(t)\}$, where $n_h$ is the count of machine settings.
- *New machine settings:* Consists of the machine settings values after they are changed at any time instant $t$ and can be represented as $h^b(t) = \{h_1^b(t), \ldots, h_{n_h}^b(t)\}$. It follows that, $h_k^b(t) = h_k(t+1)$, for any machine setting $k$ which has the same value at time $(t+1)$ as at $t$. We term the machine settings configured at production setup as **initial machine settings**, represented as $h^b(t_0)$.

Based on the above, we further define the following:

- *Machine status:* Snapshot of the machine operation at any time instant $t$ and consists of the machine sensor snapshot as well as the machine settings snapshot, and hence can be represented as $m(t) = (s(t), h(t))$.

- **Process snapshot:** Refers to the tuple $a(t) = (m(t), h^b(t))$ which represents the new machine settings based on the machine status at a time $t$.

A key advantage provided by the machine snapshot data is that it enables real-time monitoring of individual machines, as opposed to separately monitoring a large number of machine sensor and machine settings data provided by existing industrial automation software.

### C. Machine sensor and machine settings states

Machine sensor and machine settings states consist of multiple combinations of machine sensor and machine settings respectively, which are estimated to have similar impact on product quality. The machine sensor and machine settings states are defined as follows:

a) **Machine sensor states:** Combinations of machine sensor and machine settings observations which are grouped based on their impact on product quality. The machine sensor states are generated from machine statuses.
b) **Machine settings states:** Combinations of new machine settings which are grouped together based on their impact on product quality. The machine settings states are generated from new machine settings.

In order to generate the states, we use historical machine snapshots along with the corresponding process and product quality data, and consists of three major steps:

1. **Classification:** In order to estimate the impact on product quality, we build a classifier model which classifies machine snapshots with regard to product quality. The generated classifier models will then be interpreted to obtain machine sensor and machine settings states. Due to this need for interpretability [26], a decision tree is the choice of classifier used in this paper.. Further, tree-based classifiers have been previously shown to detect relationships among data which cannot be identified in regression-based method [15], [16] [17], including non-linear relationships that may exist among machine data [18].
2. **Generation of machine sensor and machine settings states:** The generated classifier models are interpreted to obtain the states. The exact method of generating states described here is specific to the choice of decision tree as a classifier. A similar method was used for process discovery in health applications [27]. For this, each decision rule from the root to a leaf node of the decision tree is converted to a state. We note here that, if a different classifier is chosen, these exact steps may change.
3. **State score:** Finally, each state is characterized using a popularity and a goodness score. The popularity is a measure of how often a state appears in the dataset, whereas the goodness score measures the likelihood of a state to achieve the target quality.

Table I shows the algorithm for generation of states from a manufacturing dataset provided as input. Machine sensor states are generated if the input for machine dataset includes the machine statuses, whereas machine settings states are generated if it includes new machine settings.

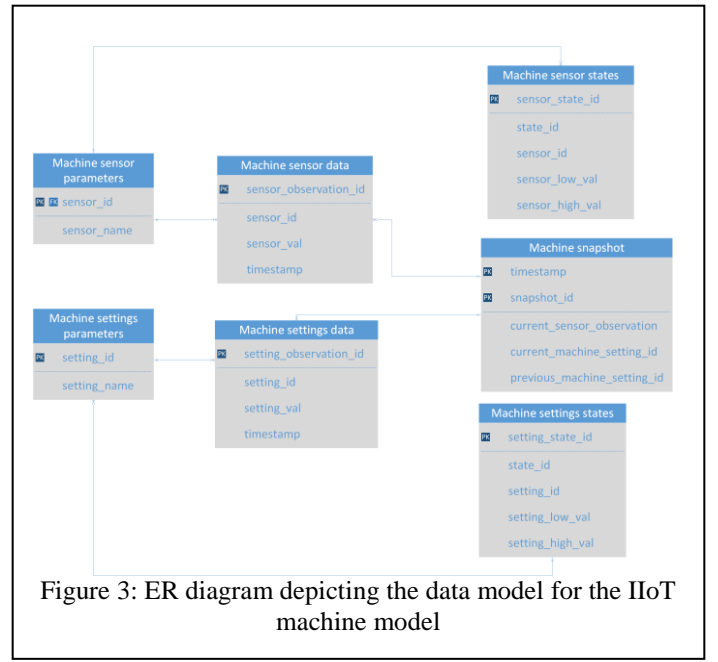

Figure 3: ER diagram depicting the data model for the IIoT machine model

TABLE I. ALGORITHM FOR GENERATION OF MACHINE SENSOR AND MACHINE SETTINGS STATES

*Input:*

Manufacturing dataset for a time period $[T_{init}, T_{final}]$, consisting of
a) Machine dataset, $V = \{v(t) : \forall t \in [T_{init}, T_{final}]\}$ where $v(t) = \{p_j, l(p_j, t), \forall p_j \in P\}$, where $P$ is the set of machine parameters and $l(p_j, t)$ is the value of the parameter $p_j$ at time $t$. Here, $V$ can either consist of machine statuses or new machine settings.
b) Production dataset consisting of data pertaining to production runs $r \in R$, such that for any production run $r$, $t_s^r \geq T_{init}, t_e^r \leq T_{final}$.
c) Quality dataset, $Q = \{q_r : r \in R, q_r \in Q_L\}$, where $q_r$ is the product quality of a production run $r$, and the target product quality is denoted as $\widehat{q_r}$

*Algorithm steps:*

**Classification**
1. Correlate machine data with production run and product quality data to obtain the training data, $X = \{x(t) = (t, v(t), r(t), q(t)) : \forall t \in [T_{init}, T_{final}]\}$. Here, $r(t)$ denotes the production run at time $t$, while $q(t)$ is the quality label for the production run $r(t)$.
2. Generate classifier model which maps machine values to quality labels, $F: V \rightarrow Q_L$

**State generation**
3. Extract a set of decision rules $\omega_i \in \Omega$ from $F$, where each rule $\omega_i = \{(p_j, o, l), \forall p_j \in P\}$ is a list of conditions with each parameter $p_j$ being associated with conditions $o \in \{>, \leq\}$ and $l \in \{p_{j,i}^{low}, p_{j,i}^{high}\}$ denoting the corresponding values for that condition.

4. Generate a set of states $C$ from $\Omega$, where each state $c_i \in C$ is given as,
$$c_i = \{(p_{1,i}^{low}, p_{1,i}^{high}), \ldots (p_{N,i}^{low}, p_{N,i}^{high})\}$$

**State scoring**

5. For each state $c_i$, the *state popularity* is measured as
$$\pi(c_i) = |x_i(t)|,$$
where $x_i(t) = \{x(t) | l(p_j, t) > p_{j,i}^{low} \;\&\; l(p_j, t) \leq p_{j,i}^{high} \;\forall\; p_j \in P\}$

6. For each state $c_i$, the *state goodness* is measured as
$$\gamma(c_i) = \frac{|x_i'(t)|}{\pi(c_i)},$$
where $x_i'(t) = \{x_i(t) | q(t) = \widehat{q_r}\}$

## V. IMPLEMENTATION OF IIOT MACHINE MODEL TO MANUFACTURING DEPLOYMENTS

In this section, we describe how the proposed IIoT machine model can be implemented in different manufacturing domains described in section II. We first outline how the IIoT machine model can be applied to two manufacturing domains, namely manufacturing of composites and steel. Subsequently, we describe in greater detail a specific manufacturing scenario of food processing, that of Vegemite production, and show, using real-world manufacturing data, how the IIoT model is adapted here.

### A. Composites manufacturing

An IIoT machine model can be used in composites manufacturing to monitor the curing process, especially to determine critical resin states at which to take action. Critical resin states can be captured using machine sensor states, and the state scores could be used to determine actions such as to increase temperature or to stop curing [28].

### B. Steel manufacturing

As steel manufacturing involves multiple machines, multiple IIoT machine models corresponding to each machine can be used to monitor each stage of manufacturing. For instance, an IIoT machine model for the furnace can be used to monitor both the temperature of the molten steel as well as the outflow of the molten steel to the caster, at which stage another IIoT machine model can be used to monitor specific aspects of casting.

### C. Food processing

Since the measurement of the quality of food products varies by the type of food product, we take the example of a specific food processing plant, namely production of Vegemite. We use real-world data from a Vegemite manufacturing facility to highlight how the proposed IIoT machine model can be adapted to Vegemite production.

*1) Yeast evaporation process for Vegemite production*

Vegemite is a popular Australian breakfast spread that is produced from leftover yeast from breweries and bakeries.

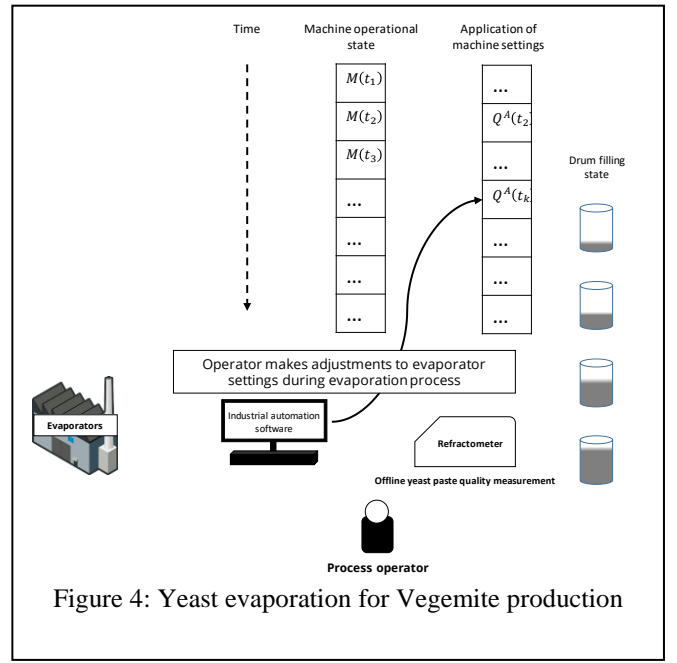

Figure 4: Yeast evaporation for Vegemite production

The yeast, which is the raw material, needs to be processed into a yeast paste, for the production of Vegemite. The yeast processing involves evaporation of the raw yeast, to meet the target quality.

The yeast evaporation process is depicted in **Error! Reference source not found.** and consists of the following steps:

1. **Raw material input:** The raw yeast undergoes pre-processing before being provided as input to the evaporators. There are three different types of raw materials, which also impact the pre-processing steps. We refer to these as *'Yeast type 1'*, *'Yeast type 2'* and *'Yeast type 3'*. The quality of the raw yeast is then tested, which is used to configure the evaporator.
2. **Production setup:** The evaporator is configured with the following at production setup:
    i. Raw material characteristics, such as the type of raw material and quality measurements, and
    ii. The initial machine settings
3. **Processing:** Once production is started with the initial machine settings, the processing stage involves monitoring of evaporators via machine sensor data and control via machine settings.
4. **Quality evaluation:** The quality evaluation of the yeast paste produced after evaporation is done offline with a refractometer, using samples periodically during production. If the measured yeast quality deviates significantly from the target quality, the yeast paste is reprocessed.

*2) Machine sensor and machine settings states of the IIoT machine model*

For the experimental evaluation, six months of manufacturing data was used to generate an IIoT machine model of the evaporator, which included data from 23 machine sensors and 7 machine settings. **Error! Reference source not found.** shows examples of two machine sensor states, with a subset of machine sensor measurements shown for readability, along with their corresponding goodness scores. For the examples shown here, we see that the two states differ in the range of values for one machine sensor, namely the `Temperature 3`, indicating that with all other values remaining the same, a higher value of 'Temperature 3' indicates that the production run is unlikely to achieve target quality.

TABLE II. EXAMPLE OF MACHINE SENSOR STATES

| Temperature 1 | Temperature 2 | Temperature 3 | Discharge solids | Density | State goodness |
|---|---|---|---|---|---|
| 18.3-86.42 | 19.1-100.61 | *16.64-66.27* | -12.99-76.42 | 0.5-1.29 | 1.0 |
| 18.3-86.42 | 19.1-100.61 | *63.90-74.34* | -12.99-76.42 | 0.5-1.29 | 0.0 |

The generated machine sensor and machine settings states can be used to perform effective monitoring and assessment of the production process. For instance, the machine sensor status and new machine settings at any time instant can be compared with the corresponding generated states to assess whether a production run is progressing as expected, based on which corrections may be made.

## VI. PREDICTIVE & PRESCRIPTIVE ANALYTICS FOR ACHIEVING CONSISTENCY IN PRODUCT QUALITY

In this section, we describe predictive and prescriptive analytics techniques that use the IIoT machine model to generate real-time product quality predictions and machine settings recommendations.

### A. Predictive analytics – real-time prediction during production

The dynamic prediction algorithm uses the instantaneous process snapshot to provide real-time predictions of product quality during a production run. These real-time predictions can be used to estimate product quality outcomes.

To do so, the machine sensor and machine settings states are combined to form composite machine sensor-setting states., As with the machine sensor and machine settings states, we obtain the popularity and goodness scores for all composite machine sensor-setting states.

The product quality prediction algorithm consists of the following steps:

I. The matching composite machine sensor-setting states for the process snapshot at time $t$ are obtained.

II. If more than one matching composite machine sensor-setting state is found, the one with the highest popularity score is chosen. For the identified composite machine sensor-setting state, the goodness score is returned as the likelihood of achieving the target product quality.

### B. Prescriptive analytics – machine settings recommendations

The recommendation algorithm provides recommendations for machine settings at a time instant $t$ given the machine status $m(t)$, in order to maximize the likelihood of achieving the target product quality.

The recommendation algorithm consists of the following steps:

I. The matching composite machine sensor-setting states for the machine status at time $t$ are obtained.

II. The composite machine sensor-setting state with the maximum goodness score is identified.

III. For the identified composite machine sensor-setting state, the machine setting state is obtained as the recommended state, with the corresponding range of values for each machine setting provided as recommended machine settings.

The table below (TABLE III) gives the algorithm for predictive and prescriptive analytics for real-time product quality prediction and machine settings recommendations to achieve target product quality.

TABLE III. PREDICTIVE AND PRESCRIPTIVE ANALYTICS ALGORITHM

***Input:***

Manufacturing dataset for a time period $[T_{init}, T_{final}]$, consisting of
a) Set of machine sensor states $SC$, and set of machine settings states $HC$
b) Process snapshot $a(t) = (m(t), h^b(t))$ at time $t$

***Algorithm steps:***

**Composite machine sensor-setting state generation**

1. The set of composite machine sensor-setting states, $D$, is generated to consist of all possible combinations of machine sensor and machine settings state. Thus, each composite machine sensor-setting state $d \in D$, is a tuple of a machine sensor state $u \in SC$ and $v \in HC$. Thus, the set $D$ is expressed as:

    $D = \{d = (u, w) | \forall u \in SC, \forall w \in HC\}$

    We use $P_D = \{P_S, P_H\}$ to denote the set of parameters in $D$, which comprises of all machine sensor parameters $P_S$ and machine settings parameters $P_H$

2. For each $d \in D$, obtain the popularity $\pi(d)$ and goodness $\gamma(d)$ scores

**Real-time product quality prediction**

3. Compute the set of matched machine sensor-settings states $D_a$ as the set of states $d \in D$ that are found to match with the process snapshot $a(t)$
4. Obtain the matched sensor-settings state $d'_a$ with highest popularity score
5. The goodness score $\gamma(d'_a)$ is obtained as the product quality prediction for the process snapshot $a(t)$

**Machine settings recommendations to achieve target product quality**

6. Compute the set of matched machine sensor-settings states $D_m$ as the set of states $d \in D$ that are found to match with the machine status $m(t)$
7. Obtain the matched sensor-settings state $d'_m$ with highest goodness score. This is the state with highest likelihood of achieving the target product quality $\widehat{q_r}$
8. For $d'_m$, the machine settings state $w'_m$ is obtained as the recommendation, with the range of values for each machine settings parameter $p \in P_H$ provided as the recommended machine settings.

## VII. EXPERIMENTAL EVALUATION OF PREDICTIVE AND PRESCRIPTIVE ANALYTICS

In this section, we present the results of applying the predictive and prescriptive analytics on the IIoT machine model generated using six months of manufacturing data from Vegemite production, described earlier in section V.C. The predictive analytics algorithms are evaluated using manufacturing data of a seventh month of Vegemite production.

### A. Predictive analytics - Real-time prediction

In order to evaluate the outcomes of predictive analytics, we compare the prediction accuracy for different configurations of the decision tree classifier model used to generate machine sensor and machine settings states. We vary the decision tree parameter "minimum leaf size" that represents the minimum number of instances in the training data set that a leaf can hold, with smaller values indicating a higher likelihood of overfitting, whereas larger values can imply more approximation [29].

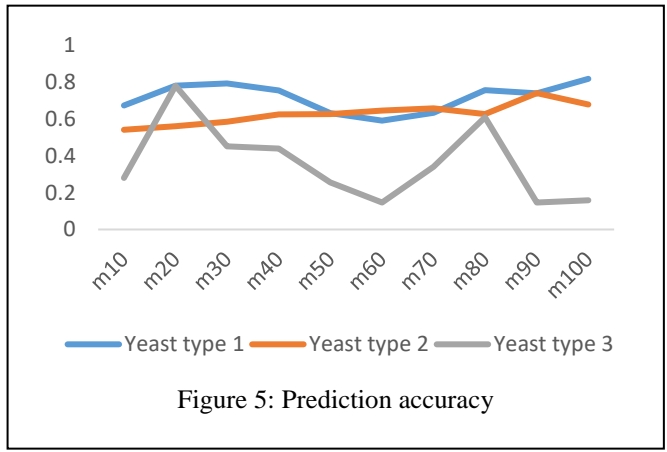

Figure 5: Prediction accuracy

We evaluate the results of product quality prediction in two ways:

1. **Prediction accuracy:** First, we look at the predictions for all process snapshots in individual production runs and compare them with the product quality outcome for the production run, and present the accuracy results as the fraction process snapshots that were correctly predicted to achieve the target quality.
2. **Correct prediction frequency:** Since the final product quality achieved depends on a series of the process snapshots, we evaluate how often correct predictions are generated for individual production runs.

Figure 5 shows how the prediction accuracy varies by the "minimum leaf size" for three types of raw materials. We see that, for 'Yeast type 1' and 'Yeast type 3', the highest prediction accuracy is achieved at a 'minimum leaf size' of 30, whereas, for 'Yeast type 2', the highest prediction accuracy is achieved for 'minimum leaf size' at 90.

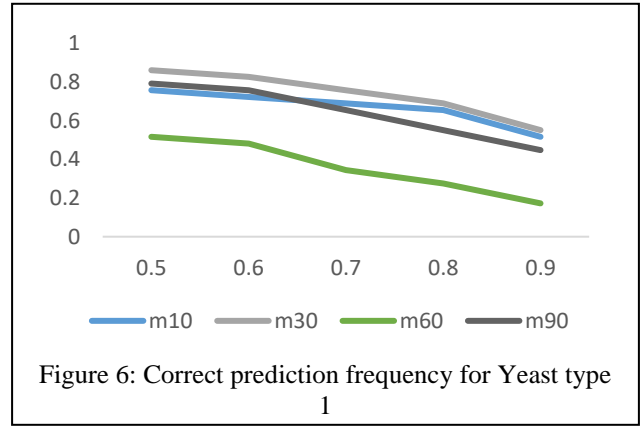

Figure 6: Correct prediction frequency for Yeast type 1

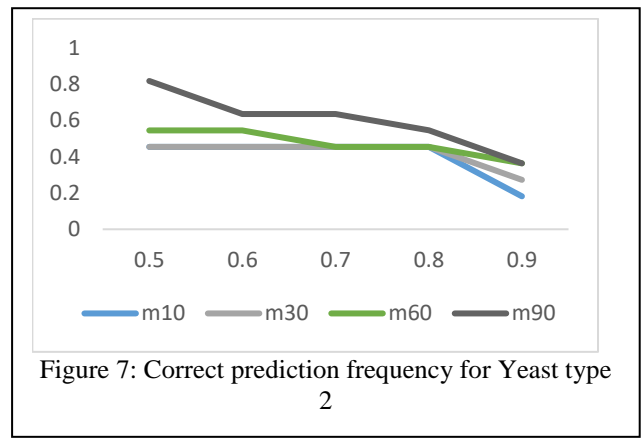

Figure 7: Correct prediction frequency for Yeast type 2

Figure 6, Figure 7 and **Error! Reference source not found.** show the results of correct prediction frequency, with the X-axis being the correct prediction frequency values, while the Y-axis is the fraction of production runs which have a correct prediction frequency higher than the corresponding value in X-axis. For 'Yeast type 1' and 'Yeast type 2', more than 90% of production runs have a correct prediction frequency higher than 0.5, thereby indicating that the real-time prediction can be used to effectively monitor production runs.

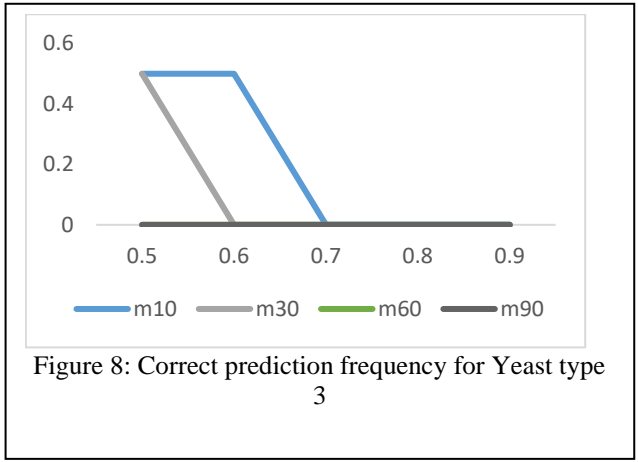

Figure 8: Correct prediction frequency for Yeast type 3

*B. Machine settings recommendations*

We present the results of machine settings recommendations obtained for specific machine sensor values. In the table below, we show how the machine settings recommendations for achieving target product quality differ from those that are unlikely to achieve the target product quality.

TABLE IV. EXAMPLES OF MACHINE SETTINGS RECOMMENDATION

| Machine sensor values | Machine settings recommendations | Machine settings not likely to achieve target quality |
|---|---|---|
| Temperature 2 19.1 – 66.5 Density 1.225 – 1.53 Production solids 2 40.6 – 101.4 | Pressure 1 104.5 – 112.5 Production solids 40.81 - 43 | Pressure 1 95.06 - 104.5 Production solids 39 – 40.56 |
| Temperature 1 47.93 – 53.07 Density 0.5 – 1.25 Level 42.48 – 103.2 | Flow setting 1482.7 – 2229.23 Pressure 1 95.06 - 104.5 Production solids 1 39 – 40.56 | Pressure 1 61 - 73.5 Production solids 1 40.85 – 43 Flow setting 2709.8 – 3278.6 |

## VIII. CONCLUSION

In this paper, we presented an IIoT machine model for effective machine monitoring and control in manufacturing environments. We also presented predictive and prescriptive analytics algorithms which use the IIoT machine model to provide decision-making inputs towards ensuring consistency of product quality. The proposed IIoT machine model and the algorithms are evaluated using a real-world manufacturing use case, and we show that the product quality can be predicted accurately using the proposed algorithms.

While the proposed IIoT machine model in this paper focuses on ensuring consistency in product quality, we note that the design of the model is quite generic, which can enable it to be adapted to other manufacturing problems.